ORIGINAL PAPER

# Discriminative Bayesian filtering lends momentum to the stochastic Newton method for minimizing log-convex functions

**Michael C. Burkhart**[1]

**Abstract**
To minimize the average of a set of log-convex functions, the stochastic Newton method iteratively updates its estimate using subsampled versions of the full objective's gradient and Hessian. We contextualize this optimization problem as sequential Bayesian inference on a latent state-space model with a discriminatively-specified observation process. Applying Bayesian filtering then yields a novel optimization algorithm that considers the entire history of gradients and Hessians when forming an update. We establish matrix-based conditions under which the effect of older observations diminishes over time, in a manner analogous to Polyak's heavy ball momentum. We illustrate various aspects of our approach with an example and review other relevant innovations for the stochastic Newton method.

**Keywords** Stochastic Newton method · Sequential Bayesian inference · Discriminative Bayesian filtering · Momentum in optimization

**Mathematics Subject Classification** 49M15 · 90C15 · 62M20 · 90C25

## 1 Optimization scheme

In machine learning and data science, we often encounter problems of the form:

$$\min_{\theta \in \mathbb{R}^d} \ell(\theta) \quad \text{for} \quad \ell(\theta) = \tfrac{1}{n} \textstyle\sum_{j=1}^{n} \log g_j(\theta) \tag{1}$$

where each log-convex function $g_j \in C^2(\mathbb{R}^d)$ corresponds to the loss accrued by an observation or sample at parameter value $\theta$ for $1 \leq j \leq n$. Examples include multiple linear regression and maximum likelihood estimation for the exponential family (see

✉ Michael C. Burkhart
mcb93@cam.ac.uk

1 University of Cambridge, Cambridge, UK

Sect. 6 for details). In this paper, we examine an online learning regime where data arrives asynchronously in a stream or $n \gg 1$ is sufficiently large that samples must be processed in batches.

We consider a sub-sampled Newton method that leverages stochastic estimates for both the gradient and Hessian [19]. At each step $t \geq 1$, we begin with the previous parameter estimate $\theta_{t-1}$, obtain a uniform random sample $\mathcal{S}_t \subset \{1, \dots, n\}$, and calculate

$$f_t = \tfrac{1}{|\mathcal{S}_t|} \textstyle\sum_{j \in \mathcal{S}_t} \nabla \log g_j(\theta_{t-1}), \tag{2a}$$

$$Q_t = \tfrac{1}{|\mathcal{S}_t|} \textstyle\sum_{j \in \mathcal{S}_t} \nabla^2 \log g_j(\theta_{t-1}), \tag{2b}$$

where $\nabla \log g_j = \nabla g_j / g_j$ and $\nabla^2 \log g_j = (g_j \nabla^2 g_j - \nabla g_j (\nabla g_j)^\intercal)/g_j^2$ denote the gradient and positive-definite Hessian of $\log g_j$, respectively. In this way, we form a step direction $-Q_t^{-1} f_t$ using only information available from the current batch $\mathcal{S}_t$. For modern applications, computer hardware limitations often constrain the batch size $|\mathcal{S}_t|$ to be much less than $n$.

Given the descent direction $-Q_t^{-1} f_t$, we perform an Armijo-style [5] backtracking line search (see Algorithm 1 for particulars) using the function $\frac{1}{|\mathcal{S}_t|} \sum_{j \in \mathcal{S}_t} \log g_j$ to determine a good step size $0 < \lambda_t < 1$ prior to updating

$$\theta_t = \theta_{t-1} - \lambda_t Q_t^{-1} f_t. \tag{3}$$

Proceeding in this way, each optimization step performs a Newton update on a sub-sampled surrogate of the true objective.

Given some initialization $\theta_0$, this method produces a sequence of estimates $\theta_1, \theta_2, \dots$ that under certain conditions tends towards the solution to problem (1). For a precise analysis of this second-order approach to stochastic optimization (in the less restrictive setting that the functions $g_j$ are convex), see Roosta-Khorasani and Mahoney [65] and Bollapragada, Byrd, and Nocedal [13].

**Thesis and outline.** Exchanging the full objective function for subsampled versions of it offers computational and practical benefits, but incurs a cost in terms of the reliability of the computed updates. In particular, the sub-sampled estimates (2a) and (2b) may prove quite noisy, hindering progress towards the optimum. This paper adapts a Bayesian filtering strategy in an attempt to mitigate this issue. We begin with a discussion of related work in the next section and then introduce discriminative Bayesian filtering in Sect. 3. We recast the optimization process described in this section as a discriminative filtering problem in Sect. 4, leading to an algorithm that calculates a step direction using the entire history of sub-sampled gradients and Hessians. In Sect. 5, we establish technical conditions under which the proposed algorithm behaves similarly to Polyak's momentum. In Sect. 6, we compare the standard approach outlined in this section to our proposed, filtered method using an online linear regression problem with synthetic data, before drawing conclusions in Sect. 7.

## 2 Related work

In this section, we provide a brief overview of related work, separated thematically into paragraphs.

*Filtering methods* have previously been applied to stochastic optimization problems, with notable success. Houlsby and Blei [33] characterized online stochastic variational inference [31] as a (non-discriminative) filtering problem using the standard Kalman filter where the covariance matrix was restricted to be isotropic and demonstrated promising results training both latent Dirichlet allocation [11, 30, 63] and Bayesian matrix factorization models [29]. For least squares problems, Bertsekas [9] demonstrated how the extended Kalman filter could be applied to form batch-based updates. More recently, Akyıldız [3] and Liu [42] developed filtered versions of the incremental proximal method [10]. In a more general setting, Stinis [70] phrased stochastic optimization as a filtering problem and proposed particle filter-based inference [77, 78].

While *momentum* and momentum-like approaches have been thoroughly explored for stochastic problems in general [14, 24, 35, 62, 66, 67, 69, 71] and for the stochastic Newton method when restricted to solving linear systems [43], momentum for more general cases of stochastic Newton has received comparatively little attention.

As the parameter space becomes high-dimensional, the computational costs for inverting the Hessian matrix grow cubically. *Hessian Free* approaches entirely circumvent the construction and subsequent inversion of the Hessian [47, 52, 75] by directly computing matrix-vector products using the conjugate gradient method or the Pearlmutter trick [58]. Berahas, Bollapragada, and Nocedal [7] explore sketching [1, 44, 55, 56, 59] as an alternative to sub-sampling.

Backtracking *line search* plays an important role in Roosta-Khorasani and Mahoney's [65] convergence results and inspired the use of line search in this work. In contrast to the more traditional stochastic approximation results that stipulate $\sum_{t=1}^{\infty} a_t = \infty$ and $\sum_{t=1}^{\infty} a_t^2 < \infty$ where $a_t > 0$ are step sizes [64], many variants of stochastic Newton use either line search or fixed step lengths. In the stochastic setting, line search remains an area of active research [8, 45, 57, 73].

Other recent innovations for the stochastic Newton method include non-uniform [76] and adaptive sampling strategies [12, 26] for the batches $\mathcal{S}_t$, low-rank approximation for the sub-sampled Hessians [25], and alternate formulations for the inverse Hessian [2]. Any of these approaches could be applied to the method we develop in the remainder of this paper.

**Algorithm 1:** Armijo-style backtracking line search

**Data:** function $h : \mathbb{R}^d \to \mathbb{R}$, initial parameter value $\theta_0$, search direction $v \in \mathbb{R}^d$, and initial gradient $m = \nabla h(\theta_0)$

**Result:** a reasonable step length $\lambda > 0$ such that $\theta_0 + \lambda v$ is nearer to $\arg\min_\theta h(\theta)$

**for** $k = 0, 1, \ldots, 4$ **do**

  $\lambda = 2^{-k}$;

  **if** $h(\theta_0 + \lambda v) - h(\theta_0) \leq 0.95\lambda \cdot \langle v, m \rangle$ **then**

    **return** $\lambda$

  **end**

**end**

**return** $\lambda$

## 3 Discriminative Bayesian filtering

Consider a state-space model relating a sequence $Z_{1:t} = Z_1, Z_2, \dots, Z_t$ of latent random variables to a corresponding sequence of observed measurements $X_{1:t} = X_1, X_2, \dots, X_t$ according to the Bayesian network:

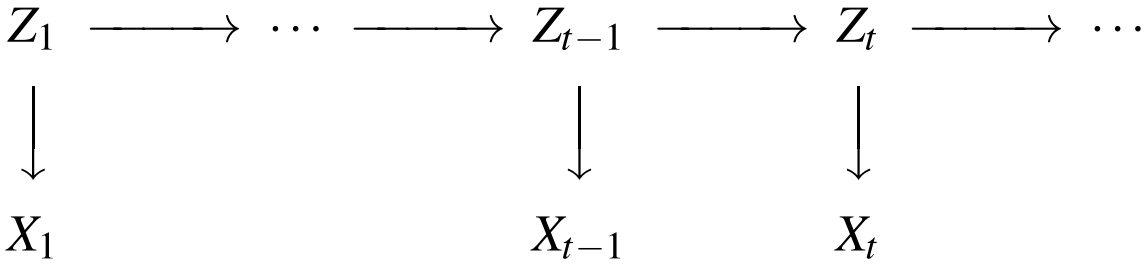

At each successive point $t$ in time, filtering aims to infer the current hidden state $Z_t$ given all currently available measurements $X_{1:t}$. We find that such an estimate often provides more accurate and more stable performance than an estimate for $Z_t$ given only the most recent measurement $X_t$. This is expected, as we know that conditioning reduces entropy [21, thm 2.6.5] and that the law of total variance[1] implies

$$\mathbb{E}[\mathbb{V}[Z_t | X_{1:t}]] \leq \mathbb{V}[Z_t | X_t]$$

where we use $\mathbb{E}[\cdot]$ and $\mathbb{V}[\cdot]$ to denote expectation and (co)variance, respectively. In particular, conditioning also reduces variance on average.

In Bayesian filtering, inference takes a distributional form. Given a state model $p(z_t | z_{t-1})$ that describes the evolution of the latent state and a measurement model $p(x_t | z_t)$ that relates the current observation and current latent state, Bayesian filtering methods iteratively infer or approximate the posterior distribution $p(z_t | x_{1:t})$ of the current latent state given all available measurements at the current point in time. To this end, the Chapman–Kolmogorov recursion

$$p(z_t | x_{1:t}) \propto p(x_t | z_t) \int p(z_t | z_{t-1}) p(z_{t-1} | x_{1:t-1}) \, dz_{t-1} \tag{4}$$

relates the current and previous posteriors in terms of the state and measurement models, up to a constant depending on the observations alone. The Kalman filter provides a quintessential example, where both the state and measurement models are chosen to be linear and Gaussian [37]. For nonlinear Gaussian models, the extended Kalman filter performs linearization prior to applying the standard Kalman updates. In general, the integrals required to compute (4) prove intractable. Assumed density filters employ variational methods to fit models to a tractable family of distributions [34, 40], sigma-point filters such as the unscented Kalman filter apply quadrature [36, 51], and particle filters perform Monte Carlo integration [27, 28]. For comprehensive surveys of Bayesian filtering, consult Chen [20] and Särkkä [68].

[1] The law of total variance states that for random variables $Y_1$ and $Y_2$ defined on the same probability space, if $\mathbb{V}[Y_1] < \infty$, then $\mathbb{V}[Y_1] = \mathbb{E}[\mathbb{V}[Y_1 | Y_2]] + \mathbb{V}[\mathbb{E}[Y_1 | Y_2]]$.

In some cases, it may be easier to calculate or approximate $p(z_t|x_t)$ than the typical observation model $p(x_t|z_t)$. In order to use the conditional distribution of latent states given observations for filtering, we may apply Bayes' rule to find that $p(x_t|z_t) \propto p(z_t|x_t)/p(z_t)$ up to a constant in $x_t$ and re-write (4) as

$$p(z_t|x_{1:t}) \propto \frac{p(z_t|x_t)}{p(z_t)} \int p(z_t|z_{t-1})p(z_{t-1}|x_{1:t-1})\, dz_{t-1}. \tag{5}$$

We characterize *discriminative filtering frameworks* as those that exchange a generative model (in the sense of Ng and Jordan [54]) for the ability to use $p(z_t|x_t)$ for inference. Well-known examples include maximum entropy Markov models [49] and conditional random fields [41], with applications including natural language processing (ibid.), gene prediction [22, 74], human motion tracking [38, 72], and neural modeling [6, 16].

In this paper, we focus on the Discriminative Kalman Filter (DKF) [17, 18] that specifies both the state and discriminative observation models as Gaussian:

$$p(z_t|z_{t-1}) = \eta_d(z_t; Az_{t-1}, \Gamma), \tag{6}$$

$$p(z_t|x_t) = \eta_d(z_t; f(x_t), Q(x_t)), \tag{7}$$

where $A \in \mathbb{R}^{d\times d}$ and $\Gamma \in \mathbb{S}_d$ parameterize the Kalman state model for the set $\mathbb{S}_d$ of valid $d\times d$ covariance matrices, $f : \mathcal{X} \to \mathbb{R}^d$ and $Q : \mathcal{X} \to \mathbb{S}_d$ parameterize the discriminative model for an abstract space $\mathcal{X}$, and $\eta_d(\cdot;\mu,\Sigma)$ denotes the $d$-dimensional Gaussian density function with mean $\mu \in \mathbb{R}^d$ and covariance $\Sigma \in \mathbb{S}_d$. With initialization $p(z_0) = \eta_d(z_0; \mathbf{0}, S)$ where $S \in \mathbb{S}_d$ satisfies $S = ASA^\intercal + \Gamma$, the unconditioned latent process is stationary. The function $f$ here may be non-linear. If the posterior at time $t-1$,

$$p(z_{t-1}|x_{1:t-1}) \approx \eta_d(z_{t-1}; \mu_{t-1}, \Sigma_{t-1}), \tag{8}$$

is approximately Gaussian then it follows from the model (6, 7) and the recursion (5) that the posterior at time $t$,

$$p(z_t|x_{1:t}) \approx \eta_d(z_t; \mu_t, \Sigma_t), \tag{9}$$

can also be approximated as Gaussian, where

$$R_{t-1} = A\Sigma_{t-1}A^\intercal + \Gamma, \tag{10a}$$

$$\Sigma_t = (Q(x_t)^{-1} + R_{t-1}^{-1} - S^{-1})^{-1}, \tag{10b}$$

$$\mu_t = \Sigma_t(Q(x_t)^{-1}f(x_t) + R_{t-1}^{-1}A\mu_{t-1}). \tag{10c}$$

In fact, this approximation is exact when the matrix $Q(x_t)^{-1} - S^{-1}$ is positive definite [18, p. 973]; if this fails to be the case, the DKF specifies $\Sigma_t = (Q(x_t)^{-1} + R_{t-1}^{-1})^{-1}$ in place of (10b). In this way, closed-form updates for the DKF's posterior require only the inversion and multiplication of $d \times d$ matrices, upon evaluation of the functions $f$ and $Q$.

In Sect. 1, we considered an optimization scheme that iteratively obtains sub-sampled values for the objective function, its gradient, and Hessian in a small neighborhood around the current parameter value. It then estimates an optimal direction of descent given only the current observations and parameter value. In this section, we showed how a discriminative Gaussian approximation (7) can be used with a latent state model (6) to consider the entire history of observations when performing inference. In the next section, we will apply this discriminative filtering process to forming updates for the stochastic Newton method.

## 4 Stochastic optimization as a filtering problem

When the batch size $|\mathcal{S}_t|$ is small, the stochastic estimates obtained for the gradient (2a) and Hessian (2b) of $\ell$ may prove to be quite noisy. To remedy this, we now outline a filtering method that incorporates multiple batches' worth of noisy measurement information to inform its estimate for $Z_t = \nabla \ell(\theta_{t-1})$. At each step, we let $X_t$ denote the current parameter value along with the function, gradient, and Hessian of $\frac{1}{|\mathcal{S}_t|} \sum_{j \in \mathcal{S}_t} \log g_j$ obtained from the uniform random sample $\mathcal{S}_t$ in a neighborhood of $\theta_{t-1}$. In order to iteratively update our distributional estimate for $Z_t$ given all available observations using the discriminative Kalman filter (DKF) as described in the previous section, we must first specify a discriminative measurement model and state model of the required form. After formulating these models, we then describe how to use the resulting filtered estimates in our optimization framework.

### 4.1 Measurement model

Given some observation $x_t$ of the random variable $X_t$, which in this case corresponds to local information for the sub-sampled function $\frac{1}{|\mathcal{S}_t|} \sum_{j \in \mathcal{S}_t} \log g_j$ in a neighborhood of $\theta_{t-1}$, we form a Gaussian approximation for the conditional distribution of $Z_t$ as

$$p(z_t|x_t) \approx \eta_d(z_t; f_t, Q_t) \tag{11}$$

where the mean $f_t$ and covariance $Q_t$ refer to (2a) and (2b), respectively. While other authors have justified similar Gaussian approximations using the sub-sampled gradient via the Central Limit Theorem [45, 46], we stress that we expect $Q_t \approx \nabla^2 \ell(\theta_t)$ in the large-sample setting: in particular, we do not intend our covariance estimate $Q_t$ to tend to zero when $d = 1$ (or toward singularity when $d > 1$).[2]

If the functions $g_j(\theta)$ in (1) are themselves probability density functions, so that we seek $\theta$ that minimizes the observed negative log likelihood

$$\log g_j(\theta) = -\log p(\theta, \psi_j) \tag{12}$$

for $\psi_1, \psi_2, \dots, \psi_n \sim^{\text{i.i.d.}} p_{\theta_*}$ and some underlying distribution $p_{\theta_*}$ where $\theta_*$ denotes the true parameter in a family $\{p_\theta\}_{\theta\in\Theta}$ of parametrized distributions, then with $\ell(\theta, \psi)$ defined analogously to (1) we have

$$\begin{aligned}\mathbb{E}[f_t] &= -\frac{1}{|\mathcal{S}_t|}\mathbb{E}\Big[\textstyle\sum_{j\in\mathcal{S}_t} \nabla_\theta \log p(\theta, \psi_j)|_{\theta=\theta_{t-1}}\Big] \\ &= -\mathbb{E}_{\Psi\sim p_{\theta_*}}[\nabla_\theta \log p(\theta, \Psi)|_{\theta=\theta_{t-1}}] = \mathbb{E}_{\Psi\sim p_{\theta_*}}[\nabla_\theta \ell(\theta, \Psi)|_{\theta=\theta_{t-1}}]\end{aligned} \tag{13}$$

so that $f_t$ from (2a) with the functions $\log g_j(\theta)$ as specified in (12) is an unbiased Monte Carlo estimate for the expected gradient of the objective. Furthermore, the Fisher information equality implies

$$\begin{aligned}\mathbb{V}_{\Psi\sim p_{\theta_*}}\left[\nabla_\theta \log p(\theta, \Psi)|_{\theta=\theta_*}\right] &= -\mathbb{E}_{\Psi\sim p_{\theta_*}}\left[\nabla^2_\theta \log p(\theta, \Psi)|_{\theta=\theta_*}\right] \\ &= \mathbb{E}_{\Psi\sim p_{\theta_*}}\left[\nabla^2_\theta \ell(\theta, \Psi)|_{\theta=\theta_*}\right]\end{aligned} \tag{14}$$

so that for $\theta_{t-1}$ near the the optimum $\theta_* = \arg\min_\theta\{\ell(\theta)\}$, we have

$$\mathbb{E}[Q_t] \approx \mathbb{V}_{\Psi\sim p_{\theta_*}}[\nabla_\theta \log p(\theta, \Psi)|_{\theta=\theta_*}] = \mathbb{V}_{\Psi\sim p_{\theta_*}}[\nabla_\theta \ell(\theta, \Psi)|_{\theta=\theta_*}] \tag{15}$$

and the sub-sampled Hessian $Q_t$ from (2b) under the specification (12) should form a reasonable approximation to the variance of the gradient. In this case, the step direction $-Q_t^{-1} f_t$ takes the form of the natural gradient [4, 48].

### 4.2 State model

We want our latent state estimate to evolve continuously, so we specify the state model

$$p(z_t|z_{t-1}) \approx \eta_d(z_t; \alpha z_{t-1}, \beta I_d) \tag{16}$$

for $0 < \alpha < 1$ and $0 < \beta$, and define $S = \frac{\beta}{1-\alpha^2} I_d$ where $I_d$ is the $d$-dimensional identity matrix. This autoregressive model with a single lag allows the previous gradient estimate to influence the current gradient estimate. In particular, we stipulate a correlation of $\alpha$ between $z_t(i)$ and $z_{t-1}(i)$ where $z(i)$ denotes the $i$-th coordinate of $z$.

[2] For mean-zero i.i.d. random vectors $Y_1, Y_2, \dots, Y_n$ with finite variance $\mathbb{V}[Y_1] = V \in \mathbb{S}_d$, the Central Limit Theorem provides conditions under which $\sqrt{n}\bar{Y}_n \to \mathcal{N}(\mathbf{0}, V)$ in distribution as $n \to \infty$, where $\bar{Y}_n$ denotes the mean of $Y_1, Y_2, \dots, Y_n$ and $\mathcal{N}(\mathbf{0}, V)$ is a $d$-dimensional Gaussian random variable with mean $\mathbf{0}$ and covariance $V$. For the unscaled mean $\bar{Y}_n$, it would follow that each component of $\mathbb{V}[\bar{Y}_n]$ tends to zero.

### 4.3 Resulting estimates and filtered optimization scheme

We now filter the state-space model described above to obtain iterative estimates for the posterior distribution. Starting with the previous approximation for the optimal descent direction given all available observations,

$$p(z_{t-1}|x_{1:t-1}) \approx \eta_d(z_{t-1};\mu_{t-1},\Sigma_{t-1}),$$

we may apply the DKF to recursively approximate the next posterior $p(z_t|x_{1:t}) \approx \eta_d(z_t;\mu_t,\Sigma_t)$ as Gaussian under (11) and (16), where

$$\Sigma_t = (Q_t^{-1} + (\alpha^2\Sigma_{t-1} + \beta I_d)^{-1} - S^{-1})^{-1}, \tag{17a}$$

$$\mu_t = \Sigma_t(Q_t^{-1}f_t + (\alpha^2\Sigma_{t-1} + \beta I_d)^{-1}\alpha\mu_{t-1}), \tag{17b}$$

if $Q_t^{-1} - S^{-1}$ is positive definite; otherwise

$$\Sigma_t = (Q_t^{-1} + (\alpha^2\Sigma_{t-1} + \beta I_d)^{-1})^{-1}.$$

This recursive approximation inspires a novel optimization scheme similar in nature to the standard stochastic Newton method introduced in Sect. 1, where we replace the unfiltered estimates $f_t$ and $Q_t$ with our filtered estimates $\mu_t$ and $\Sigma_t$, respectively, at each update step. Given the same problem (1) and initialization, at each step $t \geq 1$, we now take the search direction $-\Sigma_t^{-1}\mu_t$. We then perform an Armijo-style backtracking line search using $\frac{1}{|\mathcal{S}_t|}\sum_{j\in\mathcal{S}_t}\log g_j$. See Algorithm 2 for pseudo-code and complete details.

Calculating the posterior $p(z_t|x_{1:t})$ requires only minimal additional computational and storage costs in comparison to the standard stochastic Newton method. We introduce two hyperparameters, $\alpha$ and $\beta$, to control the influence of previous observations. Intuitively, the impact of previous updates should fade over time as our current estimate moves further away from the parameter values associated with the previously-subsampled gradients and Hessians. In the next section, we will make this intuition more precise by outlining conditions on the hyperparameters under which the impact of previous updates decays exponentially.

## 5 The connection with momentum

We would like to view our updates as analogous to Polyak's heavy ball momentum [61, 67]. In the context of optimization, momentum allows previous update directions to influence the current update direction, typically in the form of an exponentially-decaying average. This section explores how our filtered approach to optimization results in momentum-like behavior for the step direction.

To this end, we remark that from (17b) we have the recursion

$$\Sigma_t^{-1}\mu_t = Q_t^{-1}f_t + M_t\Sigma_{t-1}^{-1}\mu_{t-1}, \tag{18}$$

so that the current step direction is the sum of the current Newton update and $M_t$ times the previous step direction, where we define

$$M_t = \alpha(\alpha^2\Sigma_{t-1} + \beta I_d)^{-1}\Sigma_{t-1} \tag{19}$$

for $t \geq 2$. In the standard formulation of momentum, a scalar $0 < m < 1$ or diagonal matrix commonly takes the place of $M_t$, so that momentum acts in a coordinate-wise manner. In contrast, our matrix $M_t$ generally contains off-diagonal elements. To view our updates in the context of momentum, we need to establish matrix-based conditions for $M_t$ to dampen the impact of previous estimates over time.

For any positive-definite, Hermitian matrix $M \in \mathbb{R}^{d\times d}$, let $\lambda_{\min}(M)$ and $\lambda_{\max}(M)$ denote its smallest and largest eigenvalues, respectively. With this notation, $\rho(M) = \lambda_{\max}(M)$ corresponds to the spectral norm (as all eigenvalues are positive), and we have

**Proposition 1** *Suppose there exist* $0 < \Lambda_1 \leq \Lambda_d$ *such that* $\Lambda_1 \leq \lambda_{\min}(\Sigma_t)$ *and* $\lambda_{\max}(\Sigma_t) \leq \Lambda_d$ *for all t. If* $0 < \alpha < 1$ *and* $0 < \beta$ *are chosen to satisfy* $\alpha\Lambda_d < \alpha^2\Lambda_1 + \beta$, *then* $\rho(M_t) < 1$ *for all t.*

***Proof*** As the spectral norm is sub-multiplicative and $\lambda_{\max}(M^{-1}) = 1/\lambda_{\min}(M)$ for positive-definite matrices, we have from (19) that

$$\rho(M_t) \leq \alpha \cdot \rho\big((\alpha^2\Sigma_{t-1} + \beta I_d)^{-1}\big) \cdot \rho(\Sigma_{t-1}) \leq \alpha\Lambda_d/\lambda_{\min}(\alpha^2\Sigma_{t-1} + \beta I_d)$$

where Weyl's inequality [32, thm. 4.3.1] implies

$$\lambda_{\min}(\alpha^2\Sigma_{t-1} + \beta I_d) \geq \lambda_{\min}(\alpha^2\Sigma_{t-1}) + \lambda_{\min}(\beta I_d) \geq \alpha^2\Lambda_1 + \beta.$$

Combining the above two inequalities allows us to deduce

$$\rho(M_t) \leq \frac{\alpha\Lambda_d}{\alpha^2\Lambda_1 + \beta} < 1 \tag{20}$$

and conclude. □

We may reformulate the recursion (18) with initialization $\Sigma_1^{-1}\mu_1 = Q_1^{-1}f_1$ as

$$\Sigma_t^{-1}\mu_t = \textstyle\sum_{i=1}^{t}\big(\prod_{k=i+1}^{t} M_k\big)Q_i^{-1}f_i. \tag{21}$$

Under the conditions of the proposition, for each $i \geq 1$, we have from (20) that

$$\rho(\textstyle\prod_{k=i+1}^{t} M_k) \leq \prod_{k=i+1}^{t}\rho(M_k) \leq \big(\frac{\alpha\Lambda_d}{\alpha^2\Lambda_1+\beta}\big)^{t-i} \to 0$$

as $t \to \infty$, where

$$\left\| \left( \prod_{k=i+1}^{t} M_k \right) Q_i^{-1} f_i \right\|_2 \leq \rho\left( \prod_{k=i+1}^{t} M_k \right) \left\| Q_i^{-1} f_i \right\|_2$$

so that the impact of older updates exponentially decays over time, as one would expect from momentum.

We also note that, if $0 < \Lambda_1 \leq \Lambda_d$ exist, then $0 < \alpha < 1$ and $0 < \beta$ may always be chosen to satisfy $\alpha \Lambda_d < \alpha^2 \Lambda_1 + \beta$. For example, we may let $\alpha = 1/2$ and $\beta = \Lambda_d$.

**Algorithm 2:** Filtering for the stochastic Newton method

**Data:** initial parameter value $\theta_0$; smoothing parameters $0 < \alpha, \beta < 1$; for each random sample $\mathscr{S}_t \subset \{1, \dots, n\}$, access to the function, gradient, and Hessian of $\frac{1}{|\mathscr{S}_t|} \sum_{j \in \mathscr{S}_t} \log g_j$ from (1)

**Result:** successive parameters $\theta_1, \theta_2, \dots,$ intended to approach $\arg\min_\theta h(\theta)$

Let $S = \frac{\beta}{1-\alpha^2} I_d$;

Initialize $\mu_1 = f_1 = \frac{1}{|\mathscr{S}_1|} \sum_{j \in \mathscr{S}_1} \nabla \log g_j(\theta_0)$ and $\Sigma_1 = Q_1 = \frac{1}{|\mathscr{S}_1|} \sum_{j \in \mathscr{S}_1} \nabla^2 \log g_j(\theta_0)$;

With search direction $-\Sigma_1^{-1} \mu_1$, obtain a step size $\lambda_1$ using the Armijo-style line search from Algorithm 1 applied to $\sum_{j \in \mathscr{S}_1} \log g_j$ at $\theta_0$;

Let $\theta_1 = \theta_0 - \lambda_1 \Sigma_1^{-1} \mu_1$;

**for** $t = 2, 3, \dots$ **do**

  Let $f_t = \frac{1}{|\mathscr{S}_t|} \sum_{j \in \mathscr{S}_t} \nabla \log g_j(\theta_{t-1})$ and $Q_t = \frac{1}{|\mathscr{S}_t|} \sum_{j \in \mathscr{S}_t} \nabla^2 \log g_j(\theta_{t-1})$;

  **if** $Q_t^{-1} - S^{-1}$ *fails to be positive definite* **then**

    let $Q_t = (Q_t^{-1} + S^{-1})^{-1}$

  **end**

  Let $\Sigma_t = (Q_t^{-1} + (\alpha^2 \Sigma_{t-1} + \beta I_d)^{-1} - S^{-1})^{-1}$ and $\mu_t = \Sigma_t (Q_t^{-1} f_t + (\alpha^2 \Sigma_{t-1} + \beta I_d)^{-1} \alpha \mu_{t-1})$;

  With search direction $-\Sigma_t^{-1} \mu_t$, obtain a step size $\lambda_t$ using the Armijo-style line search from Algorithm 1 applied to $\sum_{j \in \mathscr{S}_t} \log g_j$ at $\theta_{t-1}$;

  Let $\theta_t = \theta_{t-1} - \lambda_t \Sigma_t^{-1} \mu_t$;

**end**

## 6 Illustrated example: online linear regression

In this section, we compare the *filtered method* as described in Algorithm 2 to the standard, *unfiltered method* as described in Sect. 1 on a simple optimization problem of the form (1) that we now describe.

In maximum likelihood estimation, minimizing the negative log-likelihood for a set of i.i.d. samples from a log-concave distribution produces an average of functions, each convex in the parameter of interest. We consider the problem of estimating a vector of coefficients for a discriminative linear regression model; i.e. given data $y_j \in \mathbb{R}$ and $x_j \in \mathbb{R}^d$, $1 \leq j \leq n$, we suppose

$$p_\theta(y_1, \dots y_n | x_1, \dots, x_n) = \prod_{j=1}^{n} \eta(y_j; \theta^\intercal x_j, 1), \tag{22}$$

where $\theta \in \mathbb{R}^d$ denotes a column vector of parameters. We aim to minimize the negative log likelihood, which can be written up to a multiplicative constant as

$$\ell(\theta) = \frac{1}{n}\sum_{j=1}^{n} \log g_j(\theta), \text{ where } \log g_j(\theta) = (y_j - \theta^\intercal x_j)^2/2 \tag{23}$$

so that $\nabla \log g_j(\theta) = (x_j x_j^\intercal)\theta - x_j y_j$ and $\nabla^2 \log g_j(\theta) = x_j x_j^\intercal$.

### 6.1 Methodology

We performed a computer simulation with $n = 100$ and $d = 2$. For $1 \le j \le n$, we sampled $x_j \sim^{\text{i.i.d.}} \mathcal{N}\big(\binom{0}{0}, \left(\begin{smallmatrix}1.0 & 0.1\\ 0.1 & 1.0\end{smallmatrix}\right)\big)$ and $\epsilon_j \sim^{\text{i.i.d.}} \mathcal{N}(1,1)$, where $\mathcal{N}(m, V)$ denotes a Gaussian random variable with mean $m$ and covariance $V$. We set $y_j = \theta^\intercal x_j + \epsilon_j$ for each $j$. In this way, the conditional distribution of the $y_i$ respects (22), and the minimizer of (23) corresponds to the maximum likelihood estimate (MLE) for the parameter $\theta$. Due to our choice of small $n$, the true optimum $\theta_* = \arg\min_\theta \ell(\theta)$ can be calculated exactly and used to help evaluate performance. In practical applications of stochastic Newton, we would generally expect $n$ to be much larger.

For the purposes of comparison, we ran 1000 independent paired trials starting from the same initialization. The trials performed 30 optimization steps for each method. Within each trial, the two methods received the same 5 indices $\mathcal{S}_t$, sampled uniformly at random with replacement from $\{1, \dots, n\}$ at each step. These methods then garnered gradient and Hessian information from the same subsampled function at their respective current parameter values to form each subsequent update. We selected $\alpha = 0.9$ and $\beta = 0.2$ for the filtered algorithm, but note that we generally expect these parameters to be problem-dependent.

We performed our comparisons on a 2020 MacBook Pro (Apple M1 Chip; 16 GB LPDDR4 Memory) using Python (v.3.10.2) and its Numpy package (v.1.22.3). We include code to reproduce the results and figures that follow as part of our supplementary material.

### 6.2 Results

We plot the evolution of three randomly selected paths for both methods in Fig. 1 and present a graphical summary of the aggregate results of 1000 independent trials in Fig. 2. We note that the filtered method tends to reach a neighborhood of the optimum in around 10 steps, while the unfiltered method commonly takes 30 steps or more (see Fig. 2a).

Prior to reaching a neighborhood of the optimum (where, according to Fig. 2b the function $\ell$ seems to flatten out), the filtered estimates appear to be smoother than those of the unfiltered estimates (see Fig. 1). We make this observation more numerically precise by considering the signed angular difference (in radians) between the optimal descent direction and the calculated step direction before and after the filtering process. We record the mean square of this angular error (MSE) in Table 1 for the crucial first few iterations, where both methods are taking their largest steps, and find that filtering helps to reduce MSE appreciably. As the squared bias tends to be small ($\lesssim 0.010$ for both methods over the first 5 steps), we

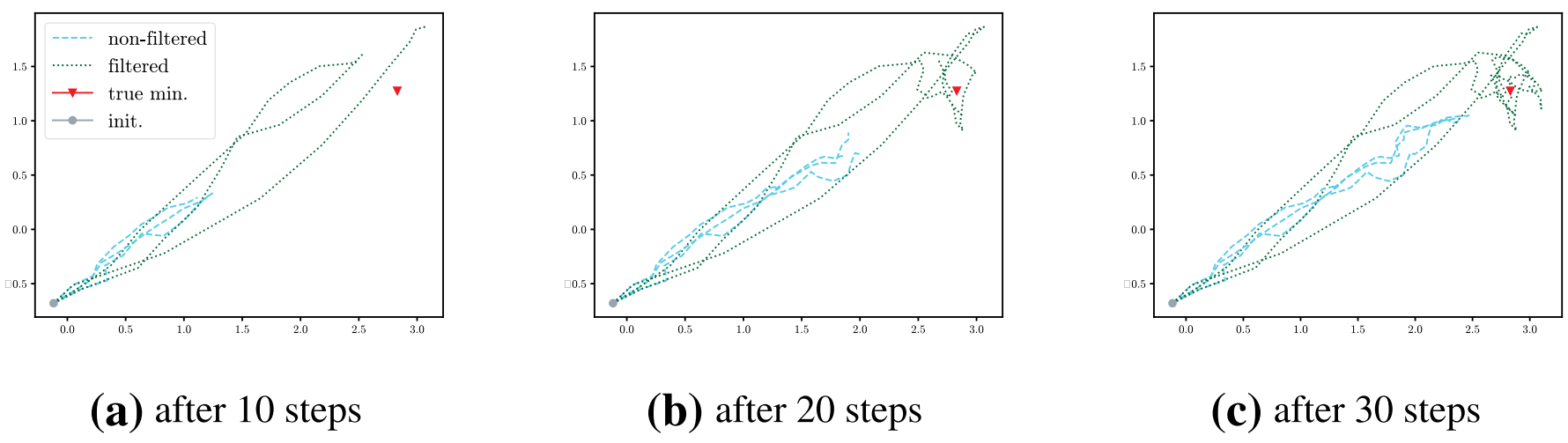

**(a)** after 10 steps **(b)** after 20 steps **(c)** after 30 steps

**Fig. 1** We plot three trajectories for the unfiltered and filtered methods (acting on the same samples) starting from the grey dot, and heading towards the global optimum (the red triangle) for the full objective function $\ell$ in (1)

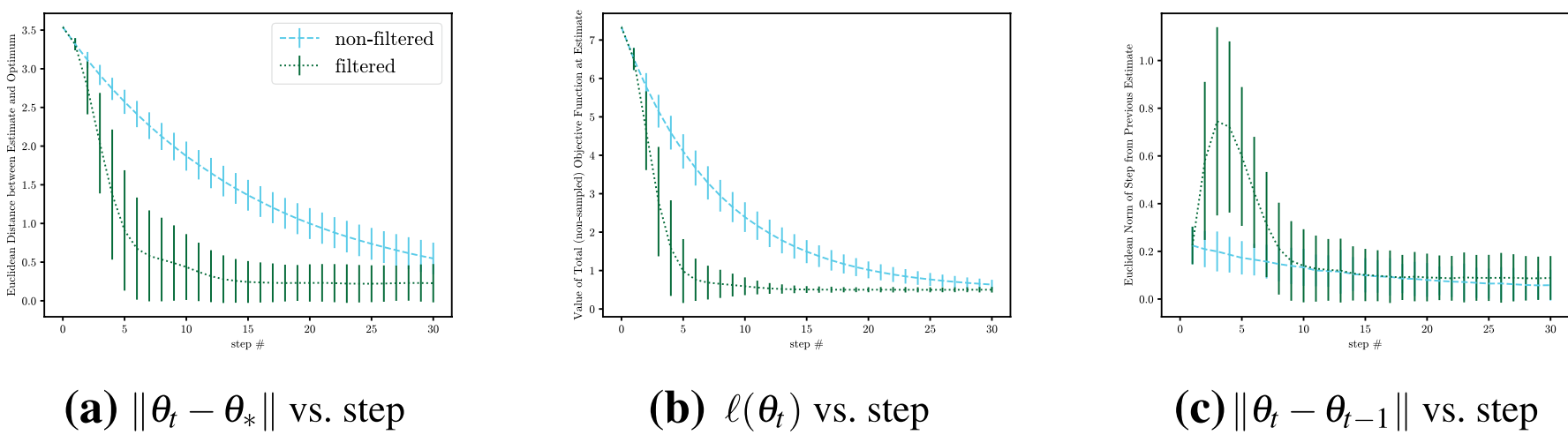

**(a)** $\|\theta_t - \theta_*\|$ vs. step **(b)** $\ell(\theta_t)$ vs. step **(c)** $\|\theta_t - \theta_{t-1}\|$ vs. step

**Fig. 2** Upon sampling 1000 trajectories for both the filtered and unfiltered method starting with the same initialization and receiving the same randomness, we plot the average values ± 2 standard deviations for **a** the Euclidean distance between the estimate and optimum, **b** the value of the entire (non-sampled) function $\ell$ at the current estimate, and **c** the distance between the current and previous estimate, all versus the step number

see a corresponding reduction in the variance of the error as well.[3] As discussed in Sect. 3, this reduction in error was one of the original motivations for applying filtering.

We note that filtering allows paths to accelerate early on in their trajectory (see Fig. 2c) and reach a neighborhood of the optimum well before the unfiltered method (see Fig. 2a). Additionally, we monitored $\rho(M_t)$, as discussed in the previous section, and found that for $t > 5$, $\rho(M_t) < 0.8$ for all 1000 trajectories. Consequently, we observe the exponential decay of the coefficients for each $Q_i^{-1} f_i$ as written in (21).

### 6.3 Exponential families and generalized linear models

We now consider how this section's linear regression example may be generalized. To this end, we introduce the exponential family [23, 39, 60], consisting of probability distributions of the form

$$p(x|\theta) = h(x) \exp(\langle \theta, T(x) \rangle - A(\theta)) \tag{24}$$

[3] For an estimator $\hat{\theta}$ of a random variable $\theta$, the mean square error decomposes as $\mathbb{E}|\theta - \hat{\theta}|^2 = \mathrm{Bias}(\hat{\theta})^2 + \mathbb{V}[\hat{\theta}]$ where $\mathrm{Bias}(\hat{\theta}) = \mathbb{E}_\theta[\hat{\theta}] - \theta$.

**Table 1** We report the mean square angular error (over the 1000 trials) for both methods during the first 5 steps. Here, the unfiltered step direction is taken to be $-Q_t^{-1} f_t$ for $f_t$ and $Q_t$ evaluated at the current *filtered* estimate. As both methods implement line search to select step length, we believe angular error (in radians) may prove more pertinent to successful optimization than other, magnitude-influenced distances. Note that both estimates coincide at step 1

| step | 1 | 2 | 3 | 4 | 5 |
|---|---|---|---|---|---|
| MSE unfiltered | 0.041 | 0.050 | 0.081 | 0.178 | 0.408 |
| MSE filtered | 0.041 | 0.043 | 0.060 | 0.093 | 0.231 |

for $x \in \mathbb{R}^\kappa$, natural parameter $\theta \in \mathbb{R}^d$, sufficient statistic $T : \mathbb{R}^\kappa \to \mathbb{R}^d$, log-normalizer $A : \mathbb{R}^d \to \mathbb{R}$, and non-negative $h : \mathbb{R}^\kappa \to \mathbb{R}$. (Note that our notation differs from that of most texts: authors typically let $\eta$ denote the natural parameter, but we use $\theta$ to maintain the notation from previous sections.) Given i.i.d. samples $x_1, \dots, x_n$ from such a distribution, the MLE for $\theta$ can be characterized as a solution to (1) using the negative log likelihood, where

$$\log g_j(\theta) := -\log p(x_j|\theta) = A(\theta) - \log h(x_j) - \langle \theta, T(x_j) \rangle$$

with

$$\nabla \log g_j(\theta) = \nabla_\theta A(\theta) - T(x_j) = \mathbb{E}_{Y \sim p_\theta}[T(Y)] - T(x_j) \tag{25}$$

and

$$\nabla_\theta^2 \log g_j(\theta) = \nabla_\theta^2 A(\theta) = \mathbb{V}_{Y \sim p_\theta}[T(Y)]. \tag{26}$$

In particular, at each optimization step, the gradient and Hessian of each $g_j(\theta_{t-1})$ will always be the expectation and variance, respectively, of $T(Y) - T(x_j)$ where $Y \sim p_{\theta_{t-1}}$.

Generalized linear models [53] with canonical response functions model conditional distributions using the exponential family. For $y_j \in \mathbb{R}$ and $x_j \in \mathbb{R}^d$, $1 \le i \le n$, and $\theta \in \mathbb{R}^d$ we write

$$p(y_j|x_j, \theta) = h(y_j)\exp(\langle \eta_j, T(y_j)\rangle - A(\eta_j)), \qquad \text{where } \eta_j = \theta^\intercal x_j, \tag{27}$$

so that the MLE for $\theta$ again solves (1) with

$$\log g_j(\theta) := -\log p(y_j|x_j, \theta) = A(\theta^\intercal x_j) - \log h(y_j) - \langle \theta^\intercal x_j, T(y_j)\rangle. \tag{28}$$

Applying the chain rule to (25) and (26) then yields $\nabla_\theta \log g_j(\theta) = x_j(\mathbb{E}[T(Y)|x_j, \theta] - T(y_j))$ and $\nabla_\theta^2 \log g_j(\theta) = (x_j x_j^\intercal)\mathbb{V}_{Y \sim p_\theta}[T(Y)]$. Thus, our algorithm may readily be applied to find the MLE for models of the above form, with a slight perturbation to the Hessian to ensure positive definiteness.

For a more standard presentation using the overdispersed exponential family, see McCullagh and Nelder [50].

## 7 Conclusions and future directions of research

The stochastic Newton algorithm uses subsampled gradients and Hessians to iteratively approximate an optimal step direction for batch-based optimization. When the batch size is small, the errors of these subsampled estimates may hinder progress towards the minimum. In this work, we applied a Bayesian filtering method with a discriminative observation model to filter the sequences of gradients and Hessians. We established conditions for the resulting optimization algorithm to behave similarly to Polyak's momentum, allowing the impact of older updates to fade over time. We illustrated how our method improves performance on a simple example and discussed how the algorithm can be applied more generally to inference for the exponential family.

In the future, we would like to consider possible solutions to two main drawbacks of our approach as currently formulated. In many practical applications, the high dimensionality of the parameter $\theta$ causes maintaining and inverting the Hessian matrix to be prohibitively expensive. Hessian free methods and the large body of research on quasi-Newton methods [15] may offer some help here. Secondly, from a theoretical perspective, our method would benefit from algorithm termination conditions and associated convergence results. The results of Roosta-Khorasani and Mahoney [65, thm. 4] and Bollapragada, Byrd, and Nocedal [13, thm. 2.2] are most germane to our work, but further modifications would be necessary.

We believe that stochastic optimization provides a natural setting for sequential Bayesian inference and anticipate further advances in this direction.

**Acknowledgements** The author would like to thank the editor Pavlo Krokhmal and the editorial staff, anonymous reviewers, and Elizabeth Crites for their thoughtful feedback on the manuscript. He also thanks Matthew Harrison and collaborators David Brandman and Leigh Hochberg for their contributions and insights developing the discriminative Kalman filter and pioneering its use in human neural decoding, and Jérôme Darbon and Basilis Gidas for their advice and encouragement.

**Code availability** Code to reproduce the table and figures presented here is publicly available at: https://github.com/burkh4rt/Filtered-Stochastic-Newton.